\documentclass[letterpaper]{article} 
\usepackage{aaai2026}  
\usepackage{times}  
\usepackage{helvet}  
\usepackage{courier}  
\usepackage[hyphens]{url}  
\usepackage{graphicx} 
\usepackage{natbib}  
\usepackage{caption} 
\usepackage{algorithm}
\usepackage{algorithmic}
\usepackage{algorithm}
\usepackage{algorithmic}
\usepackage{amsmath}
\usepackage{amsfonts}
\usepackage{booktabs}
\usepackage{multirow}
\usepackage{graphicx}
\usepackage{adjustbox}

\usepackage{newfloat}
\usepackage{listings}
\DeclareCaptionStyle{ruled}{labelfont=normalfont,labelsep=colon,strut=off} 
\floatstyle{ruled}
\newfloat{listing}{tb}{lst}{}
\floatname{listing}{Listing}
\title{FaultDiffusion: Few-Shot Fault Time Series Generation with Diffusion Model}
\author{
    Yi Xu\textsuperscript{\rm 1},
    Zhigang Chen\textsuperscript{\rm 1,\rm 2,\rm 3,*},
    Rui Wang\textsuperscript{\rm 2},
    Yangfan Li\textsuperscript{\rm 2},
    Fengxiao Tang\textsuperscript{\rm 2,*},\\
    Ming Zhao\textsuperscript{\rm 2},
    Jiaqi Liu\textsuperscript{\rm 2}
}
\affiliations{
    \textsuperscript{\rm 1}Big Data Institute, Central South University, Changsha 410083, China \\
    \textsuperscript{\rm 2}School of Computer Science and Engineering, Central South University, Changsha 410083, China\\
    \textsuperscript{\rm 3}Hunan Provincial Key Laboratory of Philosophy and Social Sciences of Urban Smart Governance, Changsha 410083, China \\
    \textsuperscript{*}Corresponding authors.\\
    \{yixu2003, czg, 234712185, liyangfan37, tangfengxiao, meanzhao, liujiaqi\}@csu.edu.cn
}

\usepackage{bibentry}

\begin{document}

\maketitle

\begin{abstract}
In industrial equipment monitoring, fault diagnosis is critical for ensuring system reliability and enabling predictive maintenance. However, the scarcity of fault data, due to the rarity of fault events and the high cost of data annotation, significantly hinders data-driven approaches. Existing time-series generation models, optimized for abundant normal data, struggle to capture fault distributions in few-shot scenarios, producing samples that lack authenticity and diversity due to the large domain gap and high intra-class variability of faults. To address this, we propose a novel few-shot fault time-series generation framework based on diffusion models. Our approach employs a positive-negative difference adapter, leveraging pre-trained normal data distributions to model the discrepancies between normal and fault domains for accurate fault synthesis. Additionally, a diversity loss is introduced to prevent mode collapse, encouraging the generation of diverse fault samples through inter-sample difference regularization. Experimental results demonstrate that our model significantly outperforms traditional methods in authenticity and diversity, achieving state-of-the-art performance on key benchmarks.
\end{abstract}


\begin{figure}[ht!]
    \centering
    \includegraphics[width=0.47\textwidth]{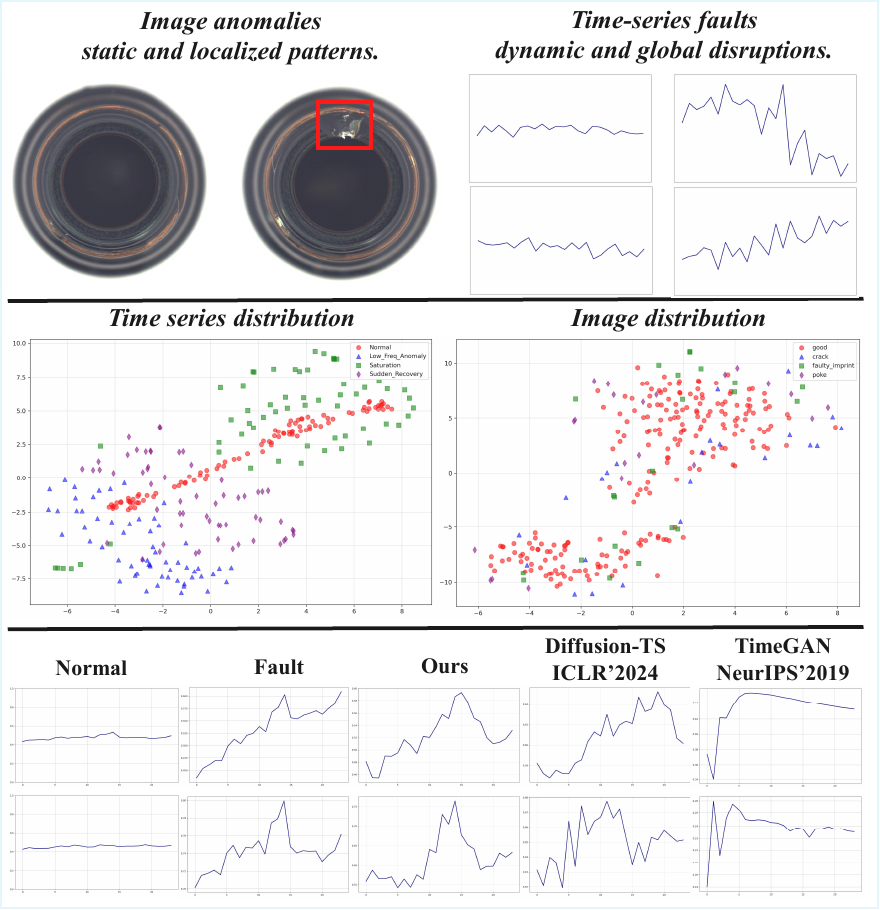}
    \caption{\textbf{Top: }Image anomalies are typically static and localized, whereas time-series faults are dynamic and global.  \textbf{Middle: } The distribution of fault time series data is significantly different from that of normal time series data, and the distribution within fault classes is diverse, while the distribution of fault images is very similar to that of normal images.\textbf{Bottom: } In experimental comparisons, our model outperformed existing anomaly generation methods in fault generation results. Compared to all other methods, our model produces the most realistic anomalies.}
    \label{intro}
\end{figure}
\section{Introduction}

In the field of industrial equipment monitoring, fault diagnosis plays a vital role in ensuring system reliability and enabling predictive maintenance\cite{Zhou2025KANAD}. However, a fundamental challenge is the extreme scarcity of fault data. This data insufficiency arises not only from the inherent rarity of fault events in production but also from the high cost and complexity in data annotation. Therefore, existing deep learning-based diagnosis methods primarily adopt an unsupervised paradigm, where one-class classifiers are trained using only abundant, fault-free data\cite{fang2024temporal,chen2024lara,li2023prototype,wang2022few}. Lacking supervision from labeled fault examples, these models, while capable of anomaly detection, cannot distinguish between different fault categories. This intrinsic limitation renders them unsuitable for crucial downstream tasks like fault classification, and hinders precise root cause analysis and targeted interventions tasks.

An intuitive solution is to synthesize additional fault samples using generative models\cite{garuti2025diffusion,ge2025t2s}. While existing time-series generation models like TimeGAN\cite{yoon2019time} and TimeVAE\cite{desai2021timevae} and CotGAN\cite{xu2020cot} and GT-GAN\cite{jeon2022gt} are effective for general data synthesis, their performance relies heavily on large training datasets\cite{jin2022domain,he2023domain}. In few-shot fault scenarios, where only a handful of examples are available, these models fail to learn the fault distribution accurately. Shown in Fig.\ref{intro} bottom, the generated samples fail to maintain the complex temporal patterns often lack authenticity compared to real-world faults.

Recently, progress in few-shot image defect generation\cite{hu2024anomalydiffusion,jin2025dual} is notable. These image-based methods operate by fine-tuning a pre-trained generator of normal data with few defect images\cite{duan2023few}, enabling pixel manipulation exclusively within well-defined, localized defect areas. However, image anomalies are typically static and localized, whereas time-series faults are dynamic events that alter the global sequence structure as illustrated in Figure\ref{intro} top. This fundamental distinction leads to two critical issues that render effective domain adaptation challenging in few-shot time-series fault settings and make the image-based methods ineffective. First, the \textbf{domain gap} between normal and fault time-series is immense. Normal data is often stable and regular, forming a stable and concentrated feature distribution. In contrast, fault data, which often contains abrupt spikes, drifts, and oscillations, exhibits a sharply divergent feature distribution. This is in contrast to image generation, where the domain shift is often smaller because the underlying context, such as object shape and texture, remains largely consistent. Second, time-series faults exhibit a remarkably \textbf{high intra-class diversity}. As illustrated in Figure\ref{intro} middle, a single type of time-series fault exhibits extensive internal diversity in its distribution, whereas image defects of the same class tend to be more uniform. This high diversity presents a severe mode collapse challenge in few-shot settings. Specifically, when a generative model is tasked with adapting to the fault domain using only a handful of highly varied fault examples, it tends to memorize and replicate these seen instances rather than learning the underlying generative rules. The result is synthetic data that, while potentially high-fidelity, fails to capture the full variety of the fault's patterns.

To address these challenges, this paper proposes a few-shot fault time series generation framework based on a diffusion model, aimed at generating high-quality and diverse fault time series data. We design a positive-negative difference adapter, which builds upon a pre-trained normal data distribution and achieves precise fault data generation by modeling the distribution differences between the normal and fault domains, effectively enabling domain adaptation. Additionally, a diversity loss is introduced to prevent degradation in the fault feature space distribution and encourage the generation of diverse fault samples through a regularization mechanism based on inter-sample differences.

To summarize, the main contributions of this paper are listed as follows:
\begin{itemize}
    \item  We propose \textbf{FaultDiffusion}, a Few-shot fault time series generation framework that efficiently generates diverse fault data from limited samples by bridging domain discrepancies. To the best of our knowledge, this is the first work on few-shot fault time series generation.
    \item  We propose a positive-negative difference adapter that leverages normal data priors to model fault distributions accurately, effectively addressing domain adaptation challenges.
    \item We propose a diversity loss that promotes sample variability and prevents degradation of the fault feature space distribution.
    \item Experiments demonstrate that our method outperforms baselines with the state-of-the-art fault time series generation quality.
\end{itemize}

\begin{figure*}[t]
    \centering
    \includegraphics[width=0.99\textwidth]{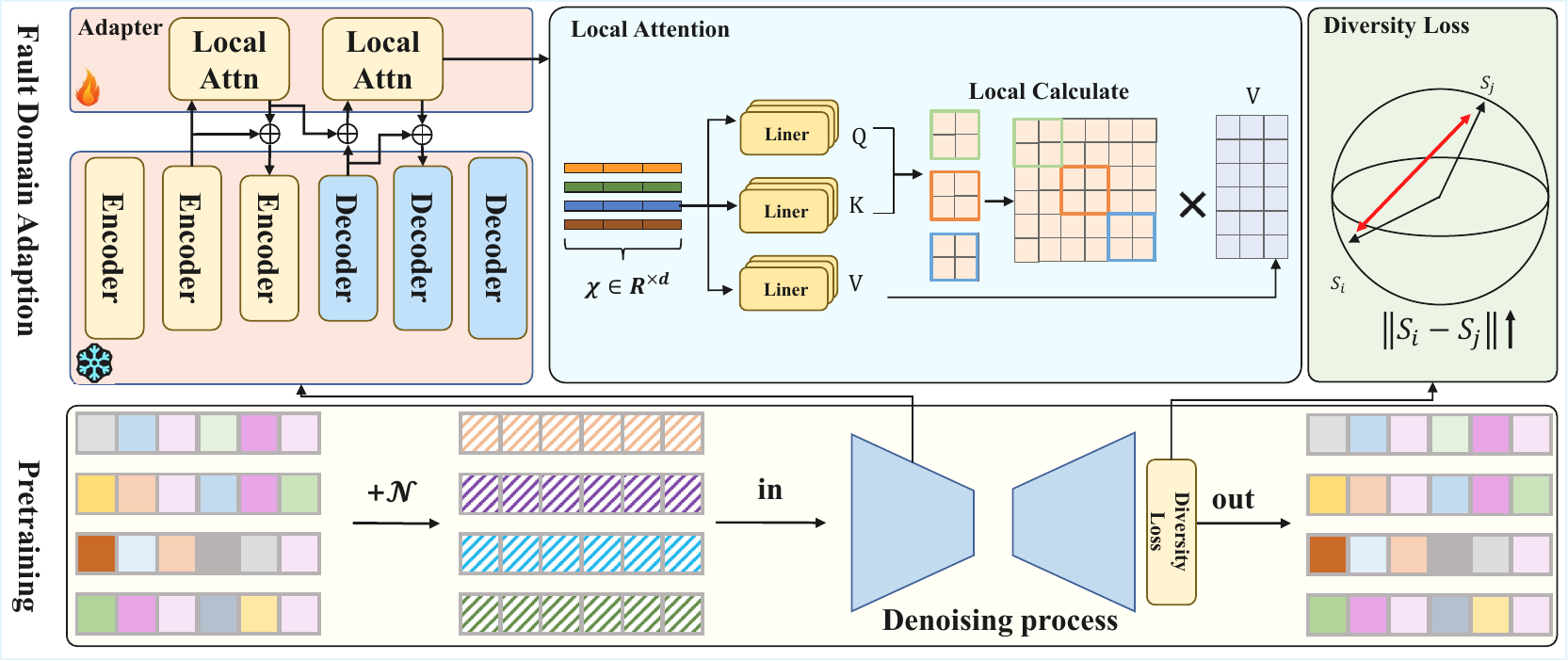}
    \caption{This is the architecture diagram of the model. The bottom part represents the pre-training process, which uses a large amount of normal data to pre-train the model to achieve normal data distribution parameter learning. The upper left corner represents the fine-tuning architecture, which achieves rapid domain adaptation through local adapters. The adapter architecture uses a local attention mechanism, as shown in the upper right figure.}
    \label{method}
\end{figure*}

\section{Method}

This section delineates the proposed methodology for few-shot fault data generation utilizing diffusion models, organized into five key components: (1) we formally define the problem of few-shot fault time series generation; (2) we detail the pretraining process, which involves training a diffusion model on normal data to capture general time-series distributions; (3) we elaborate on the transferring process, incorporating a positive-negative difference adapter for efficient migration to the fault domain and a diversity loss to enhance the variability of generated fault samples; and (4) we outline the overall training and generation procedure.
\subsection{Problem Statement}
We denote $ X_{1:\tau} = (x_1, \dots, x_\tau) \in \mathbb{R}^{\tau \times d} $ as a multivariate time series spanning $\tau$ time steps, where $d$ represents the dimension of the observed signals. Given a large dataset of normal time-series samples $ DA_n = \{ X_{1:\tau}^{n,i} \}_{i=1}^{N_n} $ with $N_n$ abundant normal instances, and a small dataset of fault time-series samples $ DA_f = \{ X_{1:\tau}^{f,j} \}_{j=1}^{N_f} $ where $N_f \ll N_n$ due to fault data scarcity.

Our primary objective is to develop a framework that first pretrains on $ DA_n $ to learn a density $ \hat p_n(x) $ that best approximates the normal data distribution $ p_n(x) $:
\begin{equation}
\min _{\hat{p_n}} D\left(\hat p_n(x) | p_n(x)\right)
\end{equation}
and then fine-tunes using $ DA_f $ to migrate from the normal domain to the fault domain for generating realistic fault time series conditioned on the normal distribution. we model the fault distribution as a conditional probability form: $ p_f(x \mid \hat{p}_n(x)) = \hat{p}_n(x) + \Delta_\theta(x), $ where $\hat{p}_n(x)$ is the pre-trained normal data distribution, $p_f(x \mid \hat p_n(x))$ is the target fault distribution conditioned on the normal distribution, and $\Delta_\theta(x)$ represents the parameterized differences learned, capturing the differences between the normal domain and the fault domain. The objective is as follow:

\begin{equation}
\min _{\hat{p_f}} D\left(\hat{p}_f(x \mid \hat p_n(x)) | p_f(x)\right)
\end{equation}

\subsection{Pretraining on Normal Data}

In the initial training phase, we aim to train a generative backbone for our framework to synthesize normal time series data by randomly sampling noise vectors. Leveraging the advanced generation capabilities of diffusion models, we adopt Diffusion-TS \cite{yuan2024diffusionts} as the backbone, which employs a transformer-based encoder-decoder architecture to process multivariate time-series data. The model breaks down the output into trend and seasonal components to capture the underlying patterns of the data.

The training process follows the Denoising Diffusion Probabilistic Model framework \cite{ho2020denoising}, involving a forward step that gradually adds noise to the data and a reverse step that learns to remove this noise. Starting from the initial time series, the forward process introduces variability, modeled by the core equation of the diffusion process:
\begin{equation}
q(x_t \mid x_{t-1}) = \mathcal{N}(x_t; \sqrt{1 - \beta_t} x_{t-1}, \beta_t I),
\end{equation}
where \( x_t \) is the noisy data at timestep \( t \), \( x_{t-1} \) is the previous clean data, \( \beta_t \) is the variance schedule, and \( I \) is the identity matrix, ensuring a controlled noise addition. The reverse process uses a transformer-based denoising network to reconstruct the original data. Pre-training is performed on a large amount of normal time series data, ensuring that the model captures a robust normal data distribution.

\subsection{Fault Domain Adaption}
To facilitate the transition from defect-free to fault time series, considering that fault time series are formed by the occurrence of faults within defect-free stable segments, we leverage the normal distribution prior from the backbone to enable migration to the fault distribution. This approach allows the framework to extend its capability to generate fault time series while preserving its ability to produce defect-free time series. Motivated by this insight, in the second training phase, we propose the integration of a positive-negative difference adapter module with the backbone, designed to generate differential representations of fault regions and their corresponding fault features. Furthermore, we introduce an additional diversity loss to supervise the fine-tuning phase, effectively preventing the degradation of the fault feature space distribution.
\subsubsection{Positive-Negative Difference Adapter}

To achieve efficient adaptation from the normal to the fault data domain, we propose a positive-negative difference adapter, which is designed to model distribution shifts using the limited fault dataset. This adapter addresses domain discrepancies by modeling the distribution shift as $ p_f(x) = p_n(x) + \Delta_\theta(x) $, where $ p_n(x) $ is the pretrained normal distribution, $ p_f(x) $ is the target fault distribution, and $ \Delta_\theta(x) $ is the learned parameterized difference. 

 Building on this, the adapter is integrated into the backbone network by freezing the encoder and decoder parameters to preserve the pretrained normal distribution, enabling the capture of anomalies and sudden changes in fault data. The adapter's output is incorporated into the backbone's residual output through a residual connection. Initially, the local input to the adapter is defined as the sum of the current backbone output and the accumulated output from previous adapters:
\begin{equation}
h_{\text{local-in}} = h_{\text{backbone-out}}^{(t)} + \sum_{k=1}^{t-1} h_{\text{local}}^{(k)},
\end{equation}
where \( h_{\text{backbone-out}}^{(t)} \) represents the backbone output at the current timestep \( t \), and \( \sum_{k=1}^{t-1} h_{\text{local}}^{(k)} \) denotes the cumulative local adapter outputs from prior timesteps, ensuring the retention of temporal context.

Subsequently, the adapter computes the local output \( h_{\text{local}}^{(t)} \) based on this input, leveraging a local attention mechanism. The updated backbone output is then refined via a residual connection:
\begin{equation}
h_{\text{backbone-out}}^{(t+1)} = h_{\text{backbone-out}}^{(t)} + \alpha h_{\text{local}}^{(t)},
\end{equation}
where \( \alpha \) is a scaling factor (typically set to 1) to control the contribution of the local adapter output, and \( h_{\text{backbone-out}}^{(t+1)} \) represents the enhanced output for the next timestep. This iterative process ensures that the adapter effectively adapts the backbone to fault-specific features while preserving the pretrained normal distribution.
The proposed interaction mechanism allows the adapter to incorporate localized fault-specific features without compromising the global representations learned by the pretrained backbone. This architectural design facilitates rapid domain adaptation with minimal trainable parameters: Only the adapter module is updated, preserving the pre-trained knowledge. Moreover, it mitigates the risk of overfitting in scenarios with limited fault samples, enhances the model’s sensitivity to anomalous patterns, and reduces overall computational complexity.

The core of the adapter lies in its sliding window attention mechanism. In order to efficiently capture local anomalies in fault data, we designed an attention mechanism based on sliding windows.
The sliding window attention within the adapter is computed as follows. Given input \( x \in \mathbb{R}^{B \times S \times D} \) (where \( B \) is the batch size, \( S \) is the sequence length, and \( D \) is the feature dimension), first pad the sequence symmetrically:
\begin{equation}
x_{\text{pad}} = \text{Pad}(x, \lfloor W/2 \rfloor),
\end{equation}
\begin{equation}
h_{\text{attn}} = \text{MultiHeadAttn}(\text{Window}(x_{\text{pad}}, W))_{\text{center}}.
\end{equation}
The sliding window attention confines focus to local neighborhoods , efficiently capturing local anomalies in fault data that global attention might overlook due to data sparsity. In few-shot fine-tuning, it reduces computational complexity, avoids overfitting, and improves generalization to fault patterns while maintaining sequence length invariance.

\begin{figure}[t]
    \centering
    \includegraphics[width=0.49\textwidth]{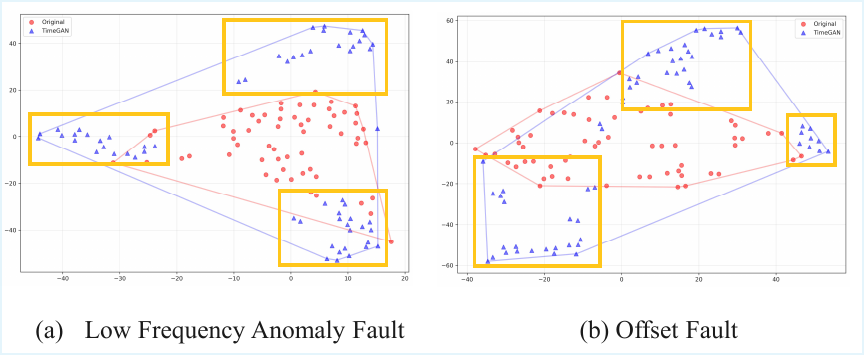}
    \caption{This is the tsne visualization of the fault samples generated by timegan. The samples are clustered and the feature space is degraded.}
    \label{loss}
\end{figure}
\subsubsection{Diversity Loss}

To address the ubiquitous mode collapse problem in fault data generation, we introduce a novel diversity loss function that aims to enhance the diversity of generated fault samples. This problem is particularly prominent in small sample scenarios, because limited fault data will aggravate model overfitting and degrade the fault feature space. As shown in the figure\ref{loss}, the feature space of fault generation samples of TimeGAN\cite{yoon2019time} is clustered together and loses diversity.

Our proposed diversity loss tackles this challenge by promoting a broader distribution of generated samples. The fine-tuning loss is composed of two key components: 
\begin{algorithm}[t]
\caption{Fault Time Series Fine-Tuning and Generation}
\label{alg:finetune}
\begin{algorithmic}
\STATE \textbf{Input}: Pretrained model \(\theta\), fault dataset \( D_{\text{fault}} \), adapters \(\phi\)
\STATE \textbf{Initialize}: Freeze encoder, initialize adapters \(\phi\), optimizer
\FOR{each fine-tuning step}
    \STATE Sample batch \( x_0 \sim D_{\text{fault}} \)
    \STATE Sample timestep \( t \sim \text{Uniform}(1, T) \)
    \STATE Compute noised data \( x_t = \sqrt{\bar{\alpha}_t} x_0 + \sqrt{1 - \bar{\alpha}_t} \epsilon \)
    \STATE Compute output \( x_{\text{out}} = \theta_{\text{enc}}(x_t, t) + \phi_{\text{dec}}(x_t, t) \)
    \STATE Compute total loss \(\mathcal{L}\)
    \STATE Update \(\phi\)
    \STATE Update Faultdiffusion model
\ENDFOR
\STATE \textbf{Generation}: Start from \( x_T \sim \mathcal{N}(0, I) \)
\FOR{\( t = T \) downto 1}
    \STATE Sample \( x_{t-1} \) using reverse process
\ENDFOR
\STATE \textbf{Output}: Generated fault samples \( x_0 \)
\end{algorithmic}
\end{algorithm}

\textbf{Base Denoising Loss}: the standard DDPM loss for accurate denoising:
  \begin{equation}
    \mathcal{L}_{\text{base}} = \mathbb{E}_{t, x_0, \epsilon} \left[ \left\| \epsilon - \epsilon_\theta(x_t, t) \right\|_1 \right].
   \end{equation}
   which measures the error between the predicted and true noise to maintain the integrity of the generated time series.
   
\textbf{Diversity Loss Design}: Enhances variety by maximizing inter-sample differences:
   \begin{equation}
    \mathcal{L}_{\text{diversity}} = \mathbb{E} \left[ \left\| s_1 - s_2 \right\|_2^2 \right], \quad s_1, s_2 \sim \epsilon_\theta(x_t, t),
  \end{equation}
   where $ s_1 $ and $ s_2 $ are randomly sampled noise predictions from the model, and the expectation is taken over these pairs to encourage diverse outputs.

The total loss is:
\begin{equation}
\mathcal{L}_{\text{total}}  = \mathcal{L}_{\text{base}} + \lambda \mathcal{L}_{\text{diversity}}.
\end{equation}
where $\lambda$ is a weighting factor that balances the trade-off between denoising accuracy and diversity, tuned to prioritize fault pattern coverage.

This diversity loss function effectively alleviates the mode collapse problem by penalizing the generation of redundant samples, thereby ensuring a wider range of fault type representations. This advantage stems from the loss function's ability to optimize the differences between samples, thereby enriching the feature space and reducing the risk of overfitting to limited fault data.

\begin{table*}[ht]
\centering
\setlength{\tabcolsep}{3.5pt}
\begin{tabular}{llcccccccccc}
\toprule
 & & \textbf{DAMF16} & \textbf{DAMF17} & \textbf{DAMF18} & \textbf{DAMF19} & \textbf{TEPF1} & \textbf{TEPF2} & \textbf{TEPF3} & \textbf{TEPF4} & \textbf{TEPF5} & \textbf{TEPF6} \\
\midrule
\multirow{4}{*}{\rotatebox[origin=c]{90}{\textbf{Cont}}} & Diffusion-TS & 69.061 & 56.444 & 96.244 & 60.074 & 36.870 & 37.408 & 39.856 & 40.982 & 33.646 & 24.833 \\
 & Cot-GAN & \textbf{58.499} & 52.212 & 120.38 & 65.632 & 43.143 & 37.150 & 45.714 & 41.784 & 40.231 & 30.498 \\
 & TimeGAN & 63.274 & 62.196 & 128.38 & 64.260 & 40.863 & 40.638 & 43.420 & 46.842 & 41.602 & 28.623 \\
 & Ours & 63.056 & \textbf{11.418} & \textbf{11.025} & \textbf{10.809} & \textbf{0.6140} & \textbf{3.3920} & \textbf{2.5030} & \textbf{2.1400} & \textbf{1.9940} & \textbf{7.2750} \\
\midrule
\multirow{4}{*}{\rotatebox[origin=c]{90}{\textbf{Corr}}} & Diffusion-TS & 203.27 & 166.07 & 178.69 & 181.87 & 267.36 & \textbf{276.37} & \textbf{78.61} & \textbf{76.64} & 224.21 & \textbf{503.58} \\
 & Cot-GAN & 233.65 & 192.81 & 221.64 & 201.73 & 237.01 & 399.77 & 282.25 & 231.06 & 208.41 & 598.51 \\
 & TimeGAN & 247.44 & 224.69 & 186.55 & 271.37 & 780.82 & 527.17 & 482.70 & 590.07 & 495.26 & 545.63 \\
 & Ours & \textbf{200.61} & \textbf{164.96} & \textbf{172.70} & \textbf{176.54} & \textbf{118.44} & 409.21 & 243.57 & 261.12 & \textbf{166.68} & 563.32 \\
\midrule
\multirow{4}{*}{\rotatebox[origin=c]{90}{\textbf{Disc}}} & Diffusion-TS & 0.5000 & 0.5000 & 0.5000 & 0.5000 & 0.4583 & 0.5000 & 0.5000 & 0.5000 & 0.5000 & 0.4917 \\
 & Cot-GAN & 0.5000 & 0.5000 & \textbf{0.4917} & 0.5000 & 0.4833 & 0.5000 & 0.5000 & 0.5000 & 0.5000 & 0.5000 \\
 & TimeGAN & 0.5000 & 0.5000 & 0.5000 & 0.4917 & 0.4917 & 0.4917 & 0.4917 & 0.5000 & 0.4583 & 0.5000 \\
 & Ours & \textbf{0.4917} & \textbf{0.4833} & 0.5000 & \textbf{0.4583} & \textbf{0.0750} & \textbf{0.3917} & \textbf{0.3833} & \textbf{0.4250} & \textbf{0.4583} & \textbf{0.4083} \\
\midrule
\multirow{4}{*}{\rotatebox[origin=c]{90}{\textbf{Pred}}} & Diffusion-TS & 0.8946 & 0.8989 & 0.8496 & 0.8564 & 1.0406 & 0.9494 & 0.9242 & 0.9202 & 0.9618 & 0.9487 \\
 & Cot-GAN & \textbf{0.8883} & 0.8811 & 0.8514 & 0.8117 & 0.9581 & 0.9301 & 0.8944 & 0.8967 & 0.9170 & 0.8960 \\
 & TimeGAN & 0.9043 & 0.8952 & 0.8515 & 0.7893 & 0.8984 & 0.8724 & 0.8996 & 0.9141 & 0.8994 & 0.8622 \\
 & Ours & 0.9132 & \textbf{0.3370} & \textbf{0.3256} & \textbf{0.3588} & \textbf{0.0668} & \textbf{0.1582} & \textbf{0.1418} & \textbf{0.1344} & \textbf{0.1238} & \textbf{0.1938} \\
\bottomrule
\end{tabular}
\caption{Results on TEP and DAMADICS datasets for generation sequence length 24. Each metric (Context FID, Correlational Mean, Discriminative Mean, Predictive Mean) is shown with methods as rows and datasets as columns. Bold indicates the best performance.}
\label{tab:comparison_all}
\end{table*}
\subsection{Training and Generation Procedure}

The training procedure consists of two phases: pretraining on normal data and fine-tuning with the adapter and diversity loss. During pretraining, the model learns general time-series distributions as described earlier. In fine-tuning, only the adapter parameters \( \phi \) are optimized, integrating positive (fault) and negative (normal) samples to model distribution differences.

The process is summarized in Algorithm~\ref{alg:finetune}. For generation, starting from pure noise \( x_T \sim \mathcal{N}(0, I) \), the reverse diffusion iteratively denoises to produce fault samples:
\[
x_{t-1} = \mu_\theta(x_t, t) + \sqrt{\Sigma_\theta(x_t, t)} \cdot z, \quad z \sim \mathcal{N}(0, I).
\]

Hyperparameters are configured to balance efficiency and performance, enabling reproducible results in few-shot scenarios.

\section{Experiment}

To validate the performance of FaultDiffusion, we conducted extensive fault generation experiments on different datasets and compared it with various methods across different metrics. The purpose of the experiments is to verify the quality of generation in the field of few-shot fault generation and the support of the generated model for downstream tasks. Our software and hardware environments are as follows: UBUNTU 18.04 LTS, PYTHON 3.8.10, PYTORCH 1.8.1, CUDA 11.2, and NVIDIA Driver 417.22, i9 CPU, and NVIDIA RTX A6000.
\subsection{Experiment Settings}
\subsubsection{Dataset}
To evaluate our few-shot fault data generation framework, we employ three time series datasets: a custom industrial dataset, the Tennessee Eastman Process (TEP) dataset, and the DAMADICS dataset. The custom dataset includes normal time series data from multi-channel sensor readings under standard operating conditions and 15 fault types---sudden fault, gradual fault, periodic fault, random noise fault, intermittent fault, impulse fault, trend fault, saturation fault, offset fault, compound fault, frequency shift fault, amplitude shift fault, missing data fault, low-frequency anomaly fault, and sudden recovery fault---designed to study small-sample fault data generation in industrial systems. The TEP dataset \cite{downs1993plant}, a benchmark for process control with 22 classes (1 normal + 21 faults, such as step changes and valve failures), from which we select six fault types for analysis. The DAMADICS dataset \cite{bartys2006damadics}, focused on valve actuators, with 20 classes (1 normal + 19 faults, including valve clogging and leakage), from which we select four fault types. These datasets are chosen for their diverse fault types and industrial relevance, enabling robust evaluation of small-sample generation and multi-class classification tasks.
\subsubsection{Mereic}
To quantitatively evaluate the quality of generated fault time series data, we employ four evaluation metrics covering distribution similarity, temporal and feature dependencies, predictive utility, and sample diversity. 1) Discriminative Score assesses distribution similarity by training a classification model to distinguish between real and synthetic data. 2) Predictive Score measures the utility of synthetic data by training a sequence prediction model on synthetic data and testing it on real data using the train-synthetic-test-real (TSTR) approach. 3) Context-Fréchet Inception Distance (Context-FID) evaluates the quality of synthetic time series by computing the difference between feature representations within local contexts. 4) Correlational Score quantifies temporal dependencies by calculating the absolute error between cross-correlation matrices of real and synthetic data.

\begin{figure*}[ht]
    \centering
    \includegraphics[width=0.99\textwidth]{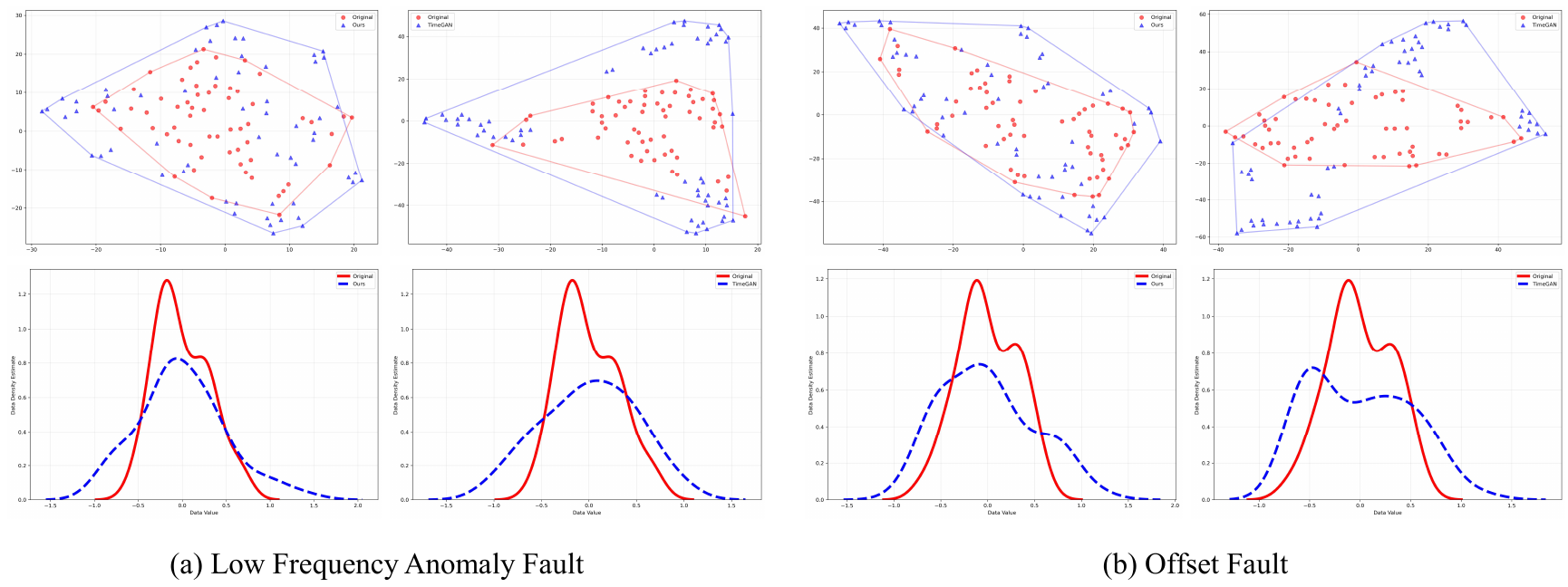}
    \caption{Visualizations of the time series synthesized by FaultDiffusion and TimeGAN.}
    \label{tsne}
\end{figure*}
\begin{table*}[t!]
\centering
\fontsize{9.2}{8}\selectfont
\setlength{\tabcolsep}{1.5pt}
\begin{tabular}{llccccccccccccccc}
\toprule
& & \textbf{FSF} & \textbf{IF} & \textbf{LFAF} & \textbf{GF} & \textbf{SaF} & \textbf{SuF} & \textbf{CF} & \textbf{RNF} & \textbf{TF} & \textbf{ImF} & \textbf{PF} & \textbf{SRF} & \textbf{MDF} & \textbf{ASF} & \textbf{OF} \\
\midrule
\multirow{5}{*}{\rotatebox[origin=c]{90}{\textbf{Cont}}} & Cot-GAN & 6.336 & 15.884 & 1.204 & 11.296 & 21.487 & 7.105 & 3.801 & 5.950 & 0.871 & 15.215 & 2.568 & 9.267 & 21.247 & 7.264 & 1.699 \\
& TimeGAN & 7.025 & 17.408 & 1.423 & 7.431 & 18.813 & 8.090 & 4.259 & 8.166 & 0.883 & 16.434 & 6.818 & 5.649 & 20.405 & 6.044 & 1.407 \\
& TimeVAE & \textbf{5.990} & 20.081 & 1.070 & 9.285 & 17.279 & 7.055 & 4.419 & 4.734 & 0.957 & 13.966 & 4.785 & 7.585 & 23.452 & 5.391 & 1.581 \\
& Diffusion-TS & 6.728 & 18.863 & 1.156 & 7.572 & 26.232 & 7.395 & 3.103 & 5.945 & \textbf{0.495} & 13.444 & 4.491 & 9.382 & 21.247 & 7.165 & 2.121 \\
& Ours & 6.081 & \textbf{12.141} & \textbf{0.961} & \textbf{4.330} & \textbf{11.025} & \textbf{3.663} & \textbf{1.645} & \textbf{2.369} & 0.804 & \textbf{9.389} & \textbf{2.105} & \textbf{3.245} & \textbf{14.483} & \textbf{2.242} & \textbf{1.131} \\
\midrule
\multirow{5}{*}{\rotatebox[origin=c]{90}{\textbf{Corr}}} & Cot-GAN & 142.72 & 91.86 & 130.73 & 122.68 & 112.30 & 124.81 & 106.45 & 115.87 & 127.21 & 128.37 & 116.03 & 128.30 & 128.64 & 119.06 & 114.75 \\
& TimeGAN & 137.12 & 131.19 & 144.26 & 151.22 & 139.05 & 121.01 & \textbf{96.28} & 139.69 & 155.01 & 132.53 & 131.69 & 132.89 & 141.80 & 133.28 & 138.68 \\
& TimeVAE & 134.21 & 90.03 & 114.80 & 118.15 & 118.75 & 115.18 & 103.04 & 118.55 & 114.78 & 110.07 & \textbf{105.49} & 115.56 & 115.83 & 112.20 & 106.71 \\
& Diffusion-TS & \textbf{117.44} & 96.29 & 120.71 & 121.50 & 109.30 & 118.26 & 107.21 & 110.29 & 120.51 & 113.52 & 127.67 & 122.78 & 125.04 & 129.07 & \textbf{92.73} \\
& Ours & 127.37 & \textbf{89.25} & \textbf{113.45} & \textbf{114.61} & \textbf{106.61} & \textbf{109.54} & 101.57 & \textbf{109.09} & \textbf{111.81} & \textbf{105.39} & 110.25 & \textbf{115.12} & \textbf{113.15} & \textbf{109.53} & 100.36 \\
\midrule
\multirow{5}{*}{\rotatebox[origin=c]{90}{\textbf{Disc}}} & Cot-GAN & 0.436 & 0.474 & 0.436 & 0.481 & 0.455 & 0.481 & 0.455 & 0.436 & 0.436 & 0.455 & 0.436 & 0.442 & 0.448 & 0.442 & 0.442 \\
& TimeGAN & 0.438 & 0.481 & 0.438 & 0.469 & 0.470 & 0.456 & 0.444 & 0.438 & 0.438 & 0.456 & 0.450 & 0.456 & 0.463 & 0.450 & 0.438 \\
& TimeVAE & 0.438 & 0.475 & 0.438 & 0.483 & 0.463 & 0.481 & 0.444 & 0.438 & 0.438 & 0.463 & 0.444 & 0.447 & 0.445 & 0.444 & 0.438 \\
& Diffusion-TS & 0.465 & 0.488 & 0.465 & 0.500 & 0.488 & 0.500 & 0.482 & 0.441 & 0.429 & 0.500 & 0.472 & 0.488 & 0.434 & 0.478 & 0.481 \\
& Ours & \textbf{0.415} & \textbf{0.382} & \textbf{0.416} & \textbf{0.440} & \textbf{0.418} & \textbf{0.443} & \textbf{0.394} & \textbf{0.418} & \textbf{0.406} & \textbf{0.394} & \textbf{0.429} & \textbf{0.441} & \textbf{0.425} & \textbf{0.429} & \textbf{0.394} \\
\midrule
\multirow{5}{*}{\rotatebox[origin=c]{90}{\textbf{Pred}}} & Cot-GAN & 0.133 & 0.323 & 0.095 & 0.239 & 0.212 & 0.284 & 0.169 & 0.108 & 0.078 & 0.232 & 0.118 & 0.149 & 0.181 & 0.155 & 0.174 \\
& TimeGAN & 0.137 & 0.338 & \textbf{0.056} & 0.255 & 0.182 & 0.147 & 0.167 & 0.090 & 0.104 & 0.242 & 0.175 & 0.124 & 0.191 & 0.088 & 0.111 \\
& TimeVAE & 0.115 & 0.336 & 0.061 & 0.216 & 0.186 & 0.258 & 0.136 & 0.091 & \textbf{0.055} & 0.221 & \textbf{0.087} & 0.149 & 0.175 & \textbf{0.087} & \textbf{0.107} \\
& Diffusion-TS & 0.135 & 0.348 & 0.067 & 0.233 & 0.246 & 0.267 & 0.148 & 0.102 & 0.064 & 0.229 & 0.135 & 0.142 & 0.171 & 0.152 & 0.145 \\
& Ours & \textbf{0.131} & \textbf{0.139} & 0.063 & \textbf{0.089} & \textbf{0.092} & \textbf{0.125} & \textbf{0.093} & \textbf{0.079} & 0.062 & \textbf{0.080} & 0.097 & \textbf{0.075} & \textbf{0.141} & 0.092 & 0.110 \\
\bottomrule
\end{tabular}
\caption{Results on Custom Industrial Dataset for Generation Sequence Length 24. Each metric (Context FID, Correlational Score, Discriminative Score, Predictive Score) is shown with methods as rows and datasets as columns. Bold indicates the best performance.}
\label{tab:comparison}
\end{table*}
\subsubsection{Baseline}
We compared the performance of our proposed model against several state-of-the-art time-series generation models for fault time series data generation, including TimeGAN\cite{yoon2019time}, Time-VAE\cite{desai2021timevae}, COT-GAN\cite{xu2020cot}, and Diffusion-TS\cite{yuan2024diffusionts}. Our approach employs a two-stage training strategy: the model is first pretrained on normal time series data with backbone weights frozen, followed by fine-tuning on few-shot fault time series data to achieve effective adaptation from the normal to the fault domain. In contrast, as the baseline methods are not designed for two-stage training, we combined normal and few-shot fault time series data for unified training in these models to ensure a fair comparison. This setup effectively validates the superiority of our model in cross-domain generation and diversity under small-sample conditions.

\subsubsection{Implementation Details}

We employ a Transformer-based Diffusion-TS model with 3 encoder and 4 decoder layers and a model dimension of 64 for few-shot fault time-series generation. Models for sequence length 24 Pre-training are trained for 25,000 epochs with a batch size of 64 and a learning rate of $1.0 \times 10^{-5}$, including 500 warm-up steps.  Fine-tuning runs for 5000 steps with a learning rate of $1.0 \times 10^{-6}$. Baselines use public code with 50,000 steps and a batch size of 128. We take the implementations and recommended hyper-parameter settings from their public codes. Each model was run five times and the average value was taken.
\subsection{Performance of Generation}

\begin{table*}[ht]
\centering

\begin{tabular}{c c c c c c}
\toprule
Adapter& Diversity Loss & Context-FID & Correlational Score & Discriminative Score & Predictive Score \\
\midrule
\checkmark & \checkmark & 5.12 & 110.45 & 0.42 & 0.10 \\
\checkmark &  & 7.35 & 120.78 & 0.45 & 0.15 \\
 & \checkmark & 8.21 & 125.63 & 0.46 & 0.18 \\
 &  & 10.47 & 130.92 & 0.48 & 0.20 \\
\bottomrule
\end{tabular}
\caption{Ablation study on our adapter and diversity loss.}
\label{abla}
\end{table*}

\begin{table}[ht!]
\centering
\setlength{\tabcolsep}{4pt}
{
\begin{tabular}{lcccc}
\toprule
\textbf{Dataset} & \textbf{Diffusion-TS} & \textbf{CotGAN} & \textbf{TimeGAN} & \textbf{Ours} \\
\cmidrule(lr){1-5}
DAMF16 & 1.3227 & 1.1995 & 1.1720 & \textbf{1.3588} \\
DAMF17 & 1.1804 & 1.2029 & 1.1055 & \textbf{1.2969} \\
DAMF18 & 1.1196 & 1.3534 & 1.1799 & \textbf{1.3701} \\
DAMF19 & 0.8186 & 1.1384 & 0.5334 & \textbf{1.2099} \\
TEPF1 & 1.8780 & 1.4227 & 0.3997 & \textbf{2.3225} \\
TEPF2 & 1.4647 & 1.6074 & \textbf{1.8545} & 1.7471 \\
TEPF3 & 1.6300 & 1.2972 & 1.3508 & \textbf{1.8318} \\
TEPF4 & 1.7004 & 1.3971 & 0.7511 & \textbf{2.1238} \\
TEPF5 & 1.7914 & 1.4224 & 1.2371 & \textbf{1.9219} \\
TEPF6 & 1.3949 & 1.5722 & 1.5144 & \textbf{1.9010} \\
\bottomrule
\end{tabular}
}
\caption{Results of TEP and DAMADICS datasets for diversity evaluation.}
\label{tab:acf_diversity_comparison}
\end{table}

\subsubsection{Fault Generation Quality}
In Table~\ref{tab:comparison}, we present the results for generating 15 types of fault time series on our custom industrial dataset, evaluating the generation quality of different methods using the aforementioned metrics. The results demonstrate that, compared to other baseline methods, our approach consistently produces higher-quality synthetic samples across nearly all metrics. Notably, our method improves the discriminative mean by approximately 20\% on average across all 15 fault types, exhibiting particularly exceptional performance in correlational score and predictive score. 
This validates the robustness of our method in tackling the challenges associated with synthesizing complex fault time series. Table~\ref{tab:comparison_all} displays the generation results on public datasets. From these results, our method achieves the best overall performance, underscoring the effectiveness of the positive-negative difference adapter and diversity loss in fault modeling. In particular, unlike other baselines, our method exhibits relatively stable performance variations across different fault types, indicating excellent robustness.To visualize the performance of time series synthesis, we adopt to
project original and synthetic data in a 2-dimensional space using t-SNE. The other is to draw data distributions using kernel density estimation. As shown in the $1^{st}$
row in Figure \ref{tsne}, FaultDiffusion overlaps original data areas markedly
better than TimeGAN. The $2^{nd}$ row in Figure \ref{tsne} shows that the synthetic data’s distributions from FaultDiffusion are more similar to those of the original data than TimeGAN.

\subsubsection{Ablation Study}
We evaluate the effectiveness of key components in our framework: adapter and diversity . Note that models without adapter rely solely on the pretrained diffusion model, while those without Diversity Loss depend on the standard denoising loss. We trained four model variants: 1) without these components; 2) with only adapter; 3) with only Diversity Loss; and 4) the full model (ours). The results for generation quality metrics are compared in Table \ref{abla}. It demonstrates that omitting any of the proposed modules leads to a noticeable decline in the model's performance on fault generation quality, which validates the efficacy of the proposed modules.

\subsubsection{Diversity Experiments}
Table \ref{tab:acf_diversity_comparison} presents the Diversity Scores for four methods across ten datasets from the TEP and DAMADICS datasets, evaluating the diversity of generated fault time series data. The results show that our method consistently achieves the highest diversity scores in nine out of ten datasets, with notable performance on TEPF1 and TEPF4 , indicating superior ability to generate varied fault patterns while maintaining similarity to real data. TimeGAN performs best on TEPF2, while Diffusionts and Cot-GAN occasionally secure second-best scores. These findings highlight our method’s robustness in producing diverse and realistic fault time series across both datasets, outperforming baseline methods in most scenarios.

\subsection{Fault generation for downstream fault diagnosis}
The experimental results for the downstream fault classification task are presented in Table \ref{down}, comparing the performance of our method against several baseline models, including CotGAN, TimeGAN, TimeVAE, and Diffusion-TS. As shown, our approach achieves the highest scores across all metrics: an accuracy of 0.8933, precision of 0.8639, recall of 0.8933, and F1-score of 0.8700. This represents an average improvement of approximately 20.5\% in accuracy and 17.3\% in F1-score over the best baseline Diffusion-TS, demonstrating the superior quality and diversity of the generated fault time series data produced by our framework.
\begin{table}[t]
\centering
\begin{tabular*}{0.48\textwidth}{l@{\extracolsep{\fill}}cccc}
\toprule
Model & Accuracy & Precision & Recall & F1-Score \\
\midrule
CotGAN & 0.6664 & 0.7279 & 0.6050 & 0.6608 \\
TimeGAN & 0.6605 & 0.6209 & 0.6605 & 0.5909 \\
TimeVAE & 0.6979 & 0.7784 & 0.6174 & 0.6886 \\
Diffusion-TS & 0.7413 & 0.7473 & 0.7353 & 0.7413 \\
Ours & \textbf{0.8933} & \textbf{0.8639} & \textbf{0.8933} & \textbf{0.8700} \\
\bottomrule
\end{tabular*}
\caption{Performance Comparison of Models on Downstream Classification Task for Time Series Generation.}
\label{down}
\end{table}

\section{Conclusion}
In this paper, we propose a few-shot fault time series generation framework, FaultDiffusion. It leverages pretraining on abundant normal data to learn general time series distributions, enabling efficient fine-tuning for fault conditions. Experiments show our method outperforms baselines, achieving state-of-the-art generation quality in authenticity, diversity, and downstream task performance, with strong robustness in unseen fault scenarios.

\bibliography{aaai2026}

\end{document}